\documentclass[sigconf, 10pt]{acmart}  

\usepackage{booktabs} 

\setcopyright{rightsretained}

\usepackage{subfigure}

\acmConference[Dr. William Cohen's ML with Large Datasets]{Machine Learning with Large Datasets}{December 2017}{CMU} 
\acmYear{2017}
\copyrightyear{2017}

\acmArticle{4}
\acmPrice{15.00}

\begin{document}
\title{Deep In-GPU Experience Replay}

\author{Ben Parr}
\affiliation{%
  \institution{Carnegie Mellon University}
}
\email{bparr@cmu.edu}

\renewcommand{\shortauthors}{B. Parr}

\begin{abstract}
Experience replay allows a reinforcement learning agent to train on samples from a large amount of the most recent experiences. A simple in-RAM experience replay stores these most recent experiences in a list in RAM, and then copies sampled batches to the GPU for training. I moved this list to the GPU, thus creating an in-GPU experience replay, and a training step that no longer has inputs copied from the CPU. I trained an agent to play Super Smash Bros. Melee, using internal game memory values as inputs and outputting controller button presses. A single state in Melee contains 27 floats, so the full experience replay fits on a single GPU. For a batch size of 128, the in-GPU experience replay trained twice as fast as the in-RAM experience replay. As far as I know, this is the first in-GPU implementation of experience replay. Finally, I note a few ideas for fitting the experience replay inside the GPU when the environment state requires more memory.

\end{abstract}

%
%



\maketitle

\section{Introduction}

Super Smash Bros.\ Melee is a video game created by Nintendo in 2001 for the GameCube that is still played competitively today. Based on ``current viewership, sponsorship, player base and, most importantly, future growth potential,'' ESPN listed Melee as number six on their 2016 Top 10 Esports Draft \cite{espn}, making Melee the longest-played game in the Esports Draft. The GameCube can be emulated on modern hardware, allowing an agent to train directly on game memory values, such a a player's exact position. In Spring 2017, Deepak Dilipkumar, Yuan Liu and I created a deep reinforcement learning agent based on Melee infrastructure code written by Vlad Firoiu. The agent learned to avoid being hit by training using these internal memory states and outputting controller button presses. The best agent learned to avoid the toughest AI built into Melee for a full minute 74.6\% of the time.

Melee allows agents to directly compete, instead of competing indirectly through a scoring system. Melee is also more complex than Atari 2600 games, and thus is harder to learn to play well. However, learning is simultaneously easier since the agent directly knows the internal memory of the game, such as characters' positions and velocities, instead of learning these features from pixels. Using internal memory values also greatly decreases the size of the memory to store an environment state. This means an experience replay containing the 1,000,000 most recent experiences can easily fit inside the GPU memory. An in-GPU experience replay reduces the communication cost between CPU and GPU because an experience is copied to the GPU memory only once, instead of once for every time it is sampled. Also, once the experience replay is in the GPU, all logic for a single train step can be moved to the GPU, thus removing even more communication between the CPU and GPU. This project achieves a significantly faster train step by reducing the amount of bytes sent between the GPU and CPU.

\begin{figure}[H]
  \centering
  \includegraphics[width=.4\textwidth]{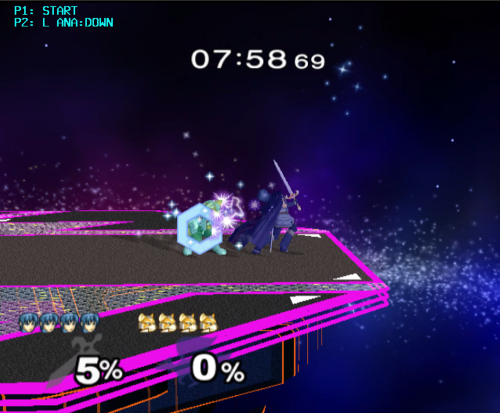}
  \caption{A screenshot of the deep reinforcement agent (Fox, left) using a special attack on the sword-wielding opponent (Marth, right) on the entirely flat Final Destination stage.}
\end{figure}

\section{Related Work}

Reinforcement learning algorithms train on experiences created by interacting with an environment \cite{sutton1998reinforcement}. An "experience" is a tuple ($s,a,r,s'$), namely an old state, an action taken, a reward for the action, and a new state (including whether the new state is a terminal state). Instead of forgetting an experience immediately after training on it, an experience replay stores a large fixed amount of the most recent experiences, with new experiences replacing old experiences in a first-in first-out order. The experience replay starts empty. Samples are initially added during the burn-in phase. Training is skipped during this burn-in phase in order to avoid sampling the initial experiences too frequently. After the burn-in phase, the agent then trains on random samples from this memory as new experiences are continually added. This random sampling smooths learning and addresses the issue of immediately forgetting a rare event \cite{lin1992self}.

DeepMind Technologies has successfully learned to play Atari 2600 games using deep reinforcement learning \cite{mnih2013playing}. DeepMind used a convolutional neural network combined with a Q-learning variant to learn to play games using just the games' pixel output, containing $84 \times 84 \times 4$ = 28,224 floats. DeepMind Technologies also scaled their Atari 2600 agent to parallel actors and parallel learners \cite{nair2015massively}. They created 100 bundles, where each bundle contained an actor that generated experiences from the Atari game, a local experience replay, and a learner that updates network parameters from the actor's experiences. Each bundle sends gradients and requests updated parameters from a distributed parameter server.

Firoiu applied deep reinforcement learning to create a Melee agent which used an actor-critic model with a deep architecture and self-play. The trained model was able to achieve excellent results with the Captain Falcon character on the Battlefield stage, going on to even beat a number of professional players \cite{firoiu2017beating}. In Spring 2017, my Deep Reinforcement Learning team trained three separate models to play Melee \cite{parr2017nintendo}. The first used DeepMind's original Deep Q-Network (DQN) architecture. The other two models were based on more recent variations of the regular DQN architecture. Double DQN \cite{double} uses two networks playing alternating roles in order to reduce the tendency of the regular DQN to over-estimate action values. Dueling DQN \cite{dueling} uses two parallel networks that estimate the value function and advantage function, and finally combine these to estimate the Q-value functions. The dueling DQN architecture performed best out of the three tested models.

The original DeepMind Atari 2600 agent, the parallelized DeepMind Atari 2600 agents and my Deep Reinforcement Learning team's agents all used a local experience replay containing 1,000,000 experiences. Open source implementations of experience replay store this large amount of experiences in RAM\footnote{DeepMind DeepMind DQN 3.0 \url{https://github.com/deepmind/dqn/blob/9d9b1d13a2b491d6ebd4d046740c511c662bbe0f/dqn/NeuralQLearner.lua\#L245}, \url{https://github.com/songrotek/DQN-Atari-Tensorflow/blob/a6320f8e585a2423a5f4e963e387c79e8fe944f0/BrainDQN_Nature.py\#L27}, \url{https://github.com/gtoubassi/dqn-atari/blob/master/replay.py\#L26} and many more.}. Therefore, every time an experience is sampled from the experience replay, it must be copied to the GPU memory. Since an experience is expected to be sampled multiple times, this creates a "thrashing" effect. My Deep Reinforcement Learning team's Melee agent was trained on a state containing only 27 floats based on internal memory values \cite{parr2017nintendo}, which means that the corresponding experience replay can fit in the memory of a single GPU. By moving the experience replay to its consumer, an experience is copied only once to the GPU memory, instead of copied every time it is sampled.

\section{Data}
Competitive Super Smash Bros. Melee is a one player versus another player fighting game where each player chooses one character to play as from the list of 25 playable characters. The goal of the game is to knock the opponent's character off the stage and into the abyss four times before the opponent knocks your character off four times. Each character has a wide array of offensive and defensive abilities. For example, the character Marth has a sword which allows him to attack from a further distance, but at a slower speed. 

Melee was originally created for the Nintendo GameCube console, which is no longer technologically advanced. In fact, Melee can be played on most modern computers using Dolphin Emulation software that emulates the GameCube hardware. Fortunately, even one of the cheapest available Google Cloud virtual machine, the g1-small machine, is able to run Dolphin-emulated Melee. This allows for easy parallelization of gameplay generation. Emulation also allows access to specific memory addresses, such as the memory address containing the position of each character. Finally, emulation allows an agent to pick the exact frame in the game to push a button. In Spring 2017, my Deep Reinforcement Learning team reused Vlad Firoiu's implementation of this infrastructure-like code \cite{firoiu2017beating}. Unfortunately this emulation suffered from sporadically skipped frames, which caused an increase in environment stochasticity.

The Melee agent is trained on a single GPU manager machine using gameplay data generated by 50 workers playing Melee in parallel. First, the manager creates an initial model, and uploads the model to the workers. The workers then enter a loop: generate a few minutes of gameplay samples from the current fixed model, upload the samples to the manager, download the latest model, repeat. Each Melee worker uses a fixed model to generate experience tuples. This allows the manager to train on a stream of incoming gameplay data generated faster than the manager could generate on its own. Finally, the manager completes the loop by periodically uploading an updated model to the workers.

\begin{figure}
\centering
\includegraphics[width=0.48\textwidth]{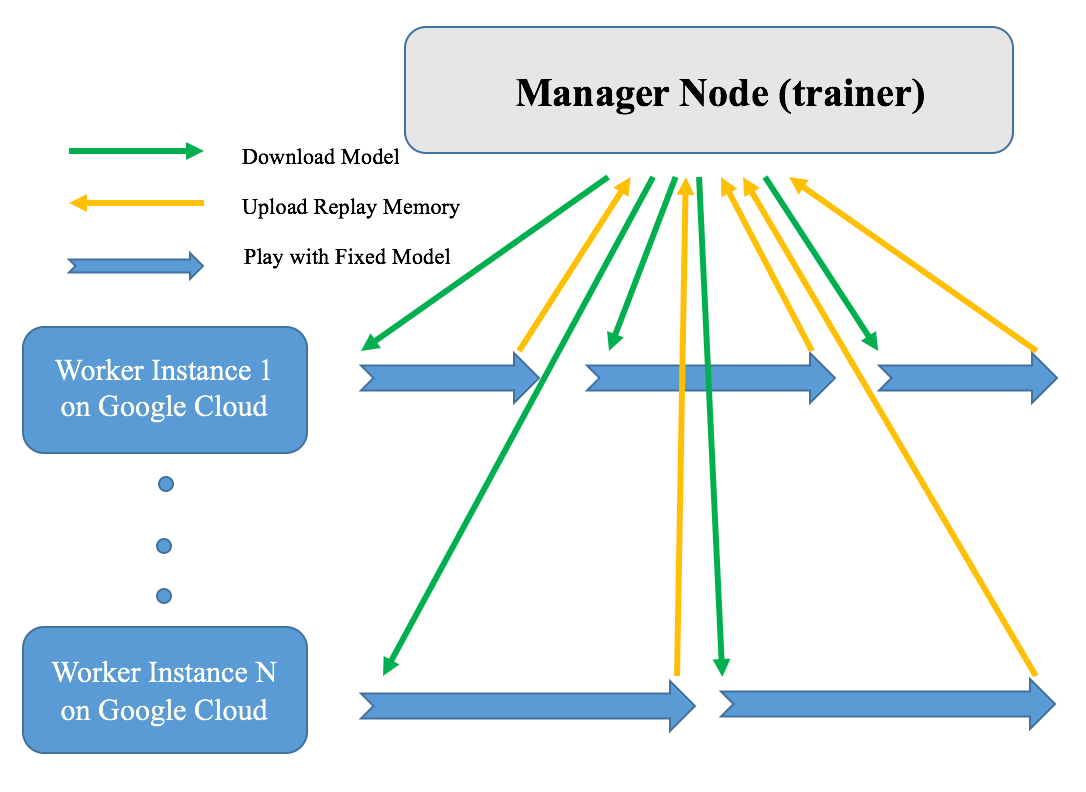}
\caption{Infrastructure outline where a single GPU-enabled manager trains on gameplay generated on N Google Cloud workers. This is the same infrastructure used in my Deep Reinforcement Learning team's Melee agent.}
\end{figure}

\section{Methods}


The in-GPU experience replay is represented in TensorFlow as a single TensorFlow Variable that stores all 1,000,000 experiences. Each experience, is packed into a single dimension. In the Melee environment, a single state consists of 27 floats. A single experience consists of two states (an old and a new state), as well as three scalars: action, reward and is\_terminal. Therefore, the experience replay Variable has shape 1,000,000 by 57. The action scalar is an integer and is\_terminal scalar is a boolean. These scalars are cast to floats when added to the experience replay, and cast back to their appropriate types when sampling. Therefore, from a user's perspective, this casting is hidden.

The Experience Replay API has two methods: adding new experiences, and randomly sampling from the current experiences. The experience replay has a fixed size and starts empty. Initially, experiences are added without removing any older experiences. Once the experience replay is completely filled, the oldest experiences are removed when newer experiences are added, in a first-in first-out order. The experience replay Variable is partitioned into contiguous blocks of equal size, which is denoted as the "update-size". Experiences are queued in RAM until the queue has enough experiences to update the next block. Then the first update-size experiences in the queue are copied to the next update block in the GPU memory using the TensorFlow scatter\_update function\footnote{tf.scatter\_update is consistently faster than tf.assign to a contiguous slice.}. This block update approach is essential since TensorFlow has a significant overhead each time values are copied to the GPU memory. Therefore, by updating a whole block at once, this overhead cost is reduced. The final update-size of 2,000 was chosen based on experimental speed results.

A sample is generated from the experience replay by first sampling integers uniformly between 0 and the current experience replay size. These integers are then used as the indexes into the experience replay Variable. Finally, the sampled experiences are unpacked into old state, new state, action, reward and is\_terminal Tensors. This approach differs slightly from the in-RAM approach which uses the \texttt{random.sample()} function, and therefore guarantees distinct experiences in a sample. The random uniform approach from the in-GPU experience replay does not have this guarantee and therefore can include a single experience multiple times in a single sample. However, since the experience replay size is much larger than the sample size, this difference happens rarely (0.05\% of the time for a sample size of 32). The overall effect of this slight difference on the trained agent's performance is unclear. Regardless, I plan on switching back to sampling distinct integers once TensorFlow has an efficient way to do so.

\subsection{Integration with Reinforcement Learning Algorithm}

Once the experience replay is located in the GPU memory, the train step logic can also be completely moved to the GPU. For example, the neural network takes as input batch-size many states and outputs batch-size many Q vectors, where each Q vector contains a Q value for each of the possible actions. The reinforcement learning algorithm then uses the sampled actions as indexes into this \texttt{network\_output}. This was originally implemented as\\ \texttt{tf.gather\_nd(network\_output, enumerate(actions))}, where \texttt{gather\_nd} selects individual scalars from a two dimensional Tensor. Unfortunately, the \texttt{enumerate} function is only available in Python, and not in TensorFlow. To convert this code to TensorFlow, I use a constant \texttt{enumerate\_mask = tf.range(0, batch\_size * num\_actions,  num\_actions)}, which contains the indexes to the start of each row of the flattened network output. Then, simply\\
\texttt{tf.gather(tf.reshape(network\_output, [-1]),}\\
\texttt{enumerate\_mask + actions)}.\footnote{This idea was taken from \url{https://github.com/tensorflow/tensorflow/issues/206}.}

Since the train step is completely moved to the GPU, there is no longer a need to feed values into the GPU when running a train step. However, for convenience, I originally allowed feed values by using \texttt{tf.placeholder\_with\_default}. This function causes a 25\% increase in train step time, which is abnormally high since the train step includes calculating gradients and updating weights! So, I removed\\
\texttt{tf.placeholder\_with\_default}, resulting in a train step with no inputs copied from the CPU.

\subsection{Deep Q-Network Model}

I used the Dueling DQN model since it performed best for my Deep Reinforcement Learning team's Melee agent. The model uses target fixing, with the target network marked as not trainable, so when the train step performs the parameter updates, only the online network is updated. The target network is instead updated to have the same weights as the online network once every 10,000 train steps. Similar with other DQN models, dueling DQN approximates the value function $Q_w$ using a deep neural network. With the weights parameterized by weights $w$, the weights are updated as:

$$ w = w + \alpha \left( r + \gamma \max_{a' \in A} Q_w(s',a') - Q_w(s,a) \right) \nabla_w Q_w(s,a)$$

The Dueling DQN architecture consists of two streams of fully-connected layers, one which estimates the state value function, and one which estimates the advantage function. The streams share one hidden layer with 128 units, and then branch out to separate fully-connected layers having 512 units. These two streams are combined to get an estimate of the Q-value for a particular (state, action) pair:

$$ Q(s,a) = V(s) + A(s,a) - \frac{1}{|\mathcal{A}|}\sum_{a'} A(s,a')$$ 

\section{Results}


Based on experimental results, I choose a GPU update-size of 2000, and saw the following percent speedups in the train step:

\begin{table}[H]
  \caption{Percent Speedup of Training from using In-GPU Experience Replay instead of In-RAM Experience Replay}
  \label{percentSpeedup}
  \begin{tabular}{ccl}
    \toprule
    Batch Size&Percent Speedup\\
    \midrule
16 & 30.54\% \\
32 & 54.87\% \\
64 & 67.08\% \\
128 & 114.41\% \\
256 & 114.35\% \\
  \bottomrule
\end{tabular}
\end{table}

All experiments were run on an NVIDIA TITAN X (Pascal) with 12 GB of memory, with no other processes running on the GPU. All results ignore startup costs, and thus begin after the burn-in phase completes. For example, creating the experience replay TensorFlow Variable takes \textasciitilde{}4 seconds. Overall, this startup time takes less than a minute for the in-GPU experience replay. Since this is a one time cost, this increased initial start-up time from using the in-GPU experience memory is negligible relative to the total training time. Finally, both the RAM implementation and GPU implementation use an experience replay size of 1,000,000, and use the same Dueling DQN model.

The RAM implementation of experience replay is able to add a new experience in 1.4 microseconds. Adding a new experience in the GPU is significantly more expensive, and depends on how many experiences are added at a time (i.e. update-size). I ran an experiment to illustrate this update-size effect, and decided on an update-size of 2000, which is able to add a new experience in 37.5 microseconds on average. Therefore, adding a new experience to the in-GPU experience replay is about 27 times slower than adding a new experience to the in-RAM experience replay.

\begin{figure}
  \centering
  \includegraphics[width=.48\textwidth]{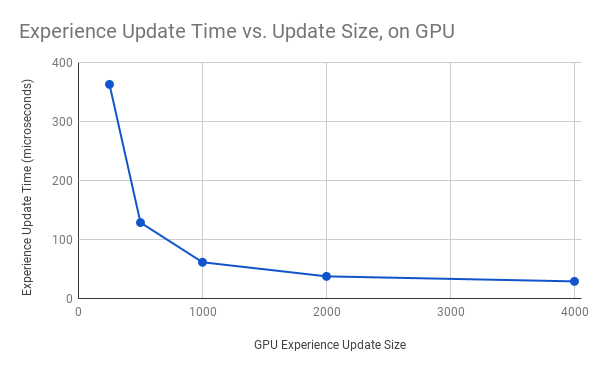}
  \caption{Average time to add a single experience when updating "GPU Experience Update Size" many experiences at a time. Lower times are better. For comparison, the corresponding update time for the in-RAM experience replay is 1.4 microseconds.}
\end{figure}

The increase in adding an experience to the in-GPU experience replay is small compared to the decrease in training time. In general, as batch size increases, the train step time should increase since the number of computed gradients also increases, for example. However, the in-RAM experience replay implementation incurs a larger slowdown as batch size increases, since it has to copy more samples to the GPU memory. The in-GPU experience replay avoids this additional cost since the samples are already in the GPU memory. Therefore, as batch size increases, the percent speedup from a in-GPU experience replay also increased. So, not only did the in-GPU experience replay train faster for all batch sizes, but also scaled better with batch size.

\begin{figure}
  \centering
  \includegraphics[width=.48\textwidth]{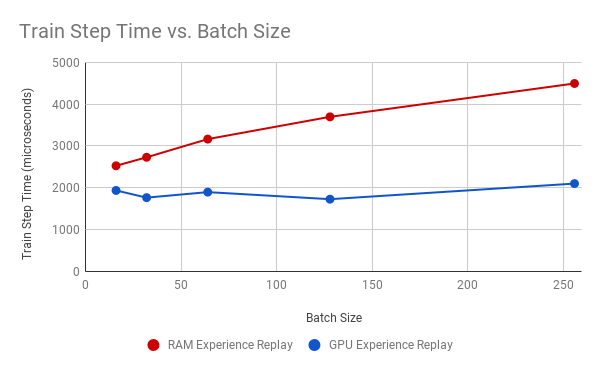}
  \caption{Average time to perform a single train step. Lower times are better. The GPU Replay Memory uses an update-size = 2000. Note that making the batch size too large could decrease the performance of the trained agent.}
\end{figure}

Finally, I compared the time it took for the Melee agent to train on 8.5 million gameplay frames. The original Melee reinforcement agent used a batch size of 32 \cite{parr2017nintendo}. For this batch size, the agent trained using the in-GPU experience replay trained 57.8\% faster than the in-RAM experience replay. Also, for batch size of 128, the in-GPU experience replay trained 104.3\% faster than the in-RAM experience replay.

\section{Discussion}

The biggest prerequisite of this project is that the the experience replay must fit inside the GPU. A Melee agent trained on internal memory values easily fits. However, an environment with pixel inputs might not. For example, DeepMind's use of game pixels stored in 28,224 floats would naively require $2 * 28224 + 3 =$ 56,451 for every experience, and thus > 50 GB of GPU memory. No existing GPU has this much memory. There are a few ways to make this tractable.

First, assuming the agent did not reach a state that\\
is\_terminal, then the new state of an experience is the old state of the next experience. So, by only storing one state per experience, and modifying the sample operations, the in-GPU experience replay could decrease the required GPU memory size by a factor of two. Also, since each gameplay added to the experience replay is a chain of temporally related experiences, there might be a way to compress the space to store a single experience. For example, the experience replay could store a combination of state and state changes, instead of always storing full states. Of course, this all increases the complexity of sampling, and reconstructing the original experience for the reinforcement learning algorithm should still all be done inside the GPU. Finally, instead of storing the original inputs as the states in the environment replay, the in-GPU experience replay could instead store the values at higher levels in the neural network as the states. Then, the layers generating these smaller states could be retrained separately and less often.

If reducing the amount of space required does not fix the issue, then the simplest solution is to just reduce the number of experiences the experience replay stores. Unfortunately this could decrease the performance of the reinforcement learning algorithm, and is not feasible if the space required is much greater than the amount of memory on a single GPU. In this case, the experience replay must be split across multiple GPUs. In order to support multi-GPU experience replay without changing training, the agent's model would have to be synchronized every train step, which could make in-RAM experience replay a better alternative. However, if willing to relax the synchronization requirement, each GPU could have its own model that is synchronized periodically, such as in DeepMind's parallel bundles approach \cite{nair2015massively}, or in an iterative parameter mixing approach \cite{mcdonald2010distributed}, or in a delayed gradient descent approach \cite{langford2009slow}.

\section{Conclusion}


I moved the experience replay from RAM to GPU memory. This restructuring removed the need to feed sampled experiences into the GPU at every train step. Instead, experiences are copied to the GPU once. Not only did this consistently speed up train time, but also decreased the incremental cost of training on larger batch sizes. For the Melee agent, this resulted in a 104.3\% faster training with a batch size of 128, and a 57.8\% faster training with a batch size of 32.

\begin{acks}
  Thank you Dr. William Cohen for feedback on this project, and for teaching Machine Learning for Large Datasets.
\end{acks}


\bibliographystyle{ACM-Reference-Format}
\nocite{*} 
\bibliography{sample-bibliography} 

\end{document}